\documentclass[letterpaper]{article} 
\usepackage{aaai2026}  
\usepackage{times}  
\usepackage{helvet}  
\usepackage{courier}  
\usepackage[hyphens]{url}  
\usepackage{graphicx} 
\usepackage{natbib}  
\usepackage{caption} 
\usepackage{algorithm}
\usepackage{algorithmic}
\usepackage{newfloat}
\usepackage{listings}
\usepackage{subfigure}
\usepackage{multirow}
\usepackage{booktabs}
\usepackage{utfsym}
\usepackage{colortbl}
\DeclareCaptionStyle{ruled}{labelfont=normalfont,labelsep=colon,strut=off} 
\floatstyle{ruled}
\newfloat{listing}{tb}{lst}{}
\floatname{listing}{Listing}
\title{U2UData+: A Scalable Swarm UAVs Autonomous Flight Dataset \\for Embodied Long-horizon Tasks}
\author{
    Tongtong Feng\textsuperscript{\rm 1}, 
    Xin Wang\textsuperscript{\rm 1}\thanks{Corresponding authors.}, 
    Feilin Han\textsuperscript{\rm 2}, 
    Leping Zhang\textsuperscript{\rm 2}, 
    Wenwu Zhu\textsuperscript{\rm 1}\footnotemark[1]
}
\affiliations{
    \textsuperscript{\rm 1}Department of Computer Science and Technology, BNRist, Tsinghua University, China\\
    \textsuperscript{\rm 2}School of Intelligent Imagery Engineering, Beijing Film Academy, China\\
    $\{$fengtongtong, xin{\_}wang, wwzhu$\}$@tsinghua.edu.cn, $\{$hanfeilin, zhangleping$\}$@bfa.edu.cn
}

\begin{document}

\maketitle

\begin{abstract}
Swarm UAV autonomous flight for Embodied Long-Horizon (ELH) tasks is crucial for advancing the low-altitude economy. However, existing methods focus only on specific basic tasks due to dataset limitations, failing in real-world deployment for ELH tasks. ELH tasks are not mere concatenations of basic tasks, requiring handling long-term dependencies, maintaining embodied persistent states, and adapting to dynamic goal shifts. This paper presents {\it U2UData+}, the first large-scale swarm UAV autonomous flight dataset for ELH tasks and the first scalable swarm UAV data online collection and algorithm closed-loop verification platform. The dataset is captured by 15 UAVs in autonomous collaborative flights for ELH tasks, comprising 12 scenes, 720 traces, 120 hours, 600 seconds per trajectory, 4.32M LiDAR frames, and 12.96M RGB frames. This dataset also includes brightness, temperature, humidity, smoke, and airflow values covering all flight routes. The platform supports the customization of simulators, UAVs, sensors, flight algorithms, formation modes, and ELH tasks. Through a visual control window, this platform allows users to collect customized datasets through one-click deployment online and to verify algorithms by closed-loop simulation. U2UData+ also introduces an ELH task for wildlife conservation and provides comprehensive benchmarks with 9 SOTA models.
\end{abstract}

\begin{links}
    \link{Dataset}{https://fengtt42.github.io/U2UData-2/}
\end{links}

\section{Introduction}
\begin{figure*}[t]
	\includegraphics[width=\textwidth]{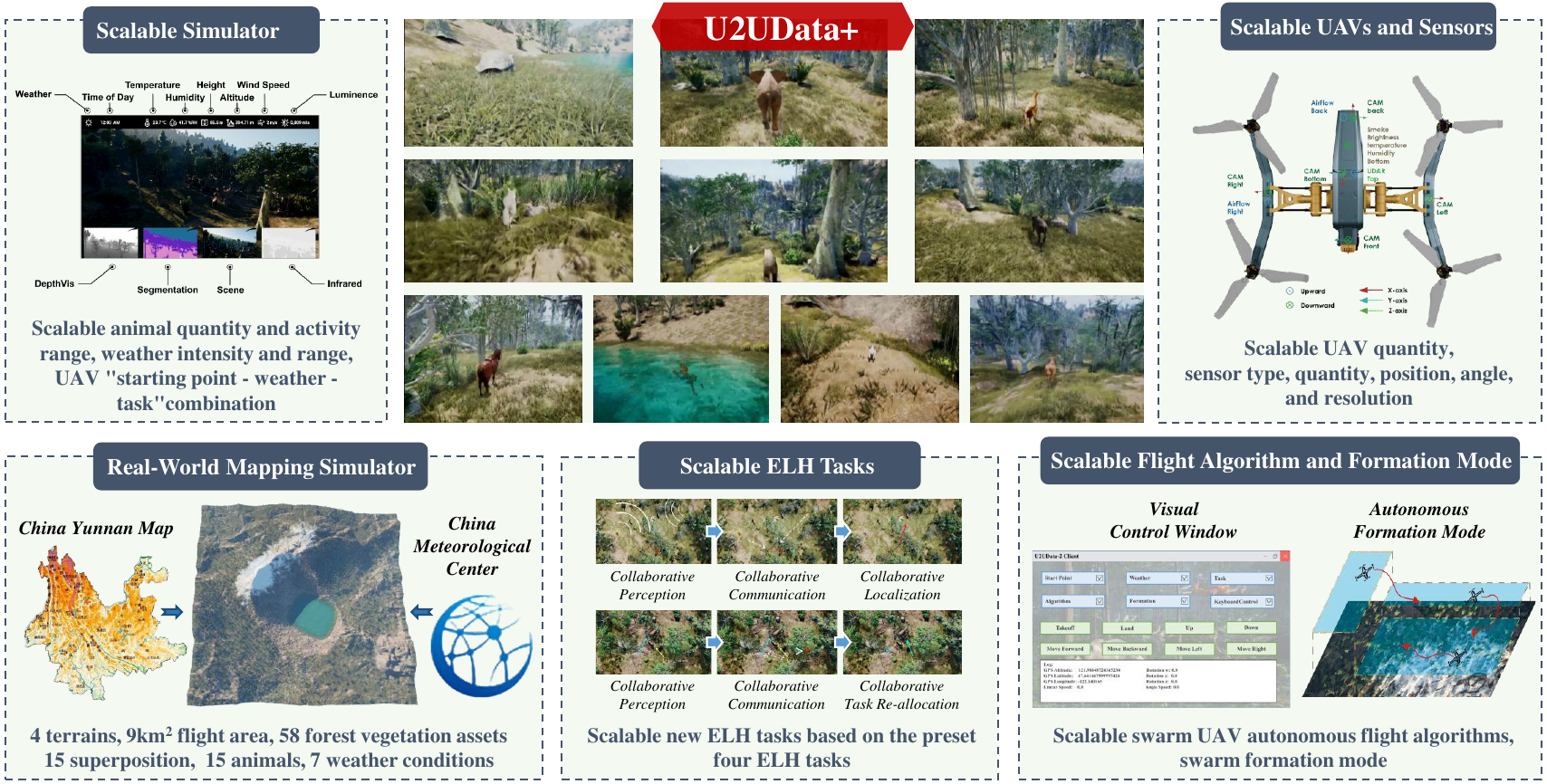}
	\centering
    \vspace{-1em}
	\caption{U2UData+ collects a large-scale swarm UAV autonomous flight dataset for ELH tasks. U2UData+ also bulid a scalable swarm UAV data online collection and algorithm closed-loop verification platform, supporting the customization of simulators, UAVs, sensors, flight algorithms, formation modes, and ELH tasks. Through a visual control window, U2UData+ allows users to collect customized datasets through one-click deployment online and to verify algorithms by closed-loop simulation.}
	\label{fig_1}
	\vspace{-0.4em}
\end{figure*}

Swarm Unmanned Aerial Vehicle (UAV) autonomous flight \cite{12} can solve the inherent limitations of single-UAV through collaborative perception, localization, communication, navigation, tracking, and task re-allocation. By leveraging UAV-to-UAV (U2U) technologies, swarm UAVs overcome single-viewpoint occlusion and sensor range limits through multi-view collaborative perception \cite{14}. Furthermore, swarm UAV ensures operational robustness against failures or obstacles for accurate navigation \cite{15} and dynamic tracking \cite{16} by collaborative localization, communication, and task re-allocation, while mitigating computational constraints via shared processing and decentralized decision-making. Finally, swarm UAV autonomous flight can achieve robust, scalable, and adaptive task execution in complex and harsh environments unattainable by single-UAV systems.

Swarm UAV autonomous flight for Embodied Long-Horizon (ELH) tasks is crucial for advancing the low-altitude economy. ELH tasks \cite{feng2025evoagent} are complex, multi-step tasks that require sustained embodied planning, sequential decision-making, and extended execution over a prolonged period to achieve a final goal. The practical applications of swarm UAV are almost all ELH tasks, such as logistics distribution \cite{betti2024uav}, wildlife conservation \cite{feng2024u2udata}, disaster rescue \cite{sun2024joint}, and infrastructure inspection \cite{pan2024unmanned}.

However, existing methods focus only on specific basic tasks due to dataset limitations, failing in real-world deployment for ELH tasks. Existing swarm UAV flight datasets, as shown in Table \ref{table1}, CoPerception-UAVs \cite{17} and CoPerception-UAVs+ \cite{36} are based on open-source simulators such as AirSim \cite{airsim} and CARLA \cite{CARLA} and consider only 1 terrain, 1 weather, and 1 to 2 sensor types; they collect datasets using fixed altitude and consistent or fixed formation mode. In real-world scenarios, compared to autonomous driving \cite{liu2024survey}, autonomous flight has more freedom, faces more complex environments, and is more susceptible to the influence of temperature, humidity, and airflow due to its smaller size. Obviously, there will be a clear domain gap between existing synthetic data and real-world data. U2UData \cite{feng2024u2udata} is the first swarm UAVs autonomous flight dataset, which is collected by 3 UAVs flying autonomously in the U2USim \cite{u2usim}, covering 4 terrains, 7 weather conditions, and 8 sensor types. Due to the emergence of U2UData, swarm UAV autonomous flight algorithms have begun to be studied. But U2UData is hard to use for exploring ELH tasks: 1) the length of each trajectory in U2UData is only 15 seconds and only focuses on basic collaborative perception and tracking tasks; 2) the dataset size, tasks, and settings are preset and fixed and cannot be expanded. ELH tasks are not mere concatenations of basic tasks, requiring handling long-term dependencies, maintaining persistent states, and adapting to dynamic goal shifts. Therefore, building a scalable swarm UAV autonomous flight dataset for ELH tasks is an urgent and challenging work for real-world deployment.

In this paper, we present {\it U2UData+}, as shown in Figure \ref{fig_1}, the first large-scale swarm UAV autonomous flight dataset for ELH tasks and the first scalable swarm UAV data online collection and algorithm closed-loop verification platform. 1) The dataset is captured by 15 UAVs in autonomous collaborative flight for ELH tasks (dataset size: 3.62T), comprising 12 scenes (weather and terrain combination), 720 traces, 120 hours (each trace 600 seconds), 4.32M LiDAR frames, 12.96M RGB, and 12.96M depth frames. This dataset also includes brightness, temperature, humidity, smoke, and airflow values covering all flight routes. 2) The platform supports the customization of simulators, UAVs, sensors, flight algorithms, formation modes, and ELH tasks. Through a visual control window, this platform allows users to collect customized datasets through one-click deployment online and to verify algorithms by closed-loop simulation, which can greatly alleviate the limitations of existing datasets on algorithm development. 3) U2UData+ also introduces an ELH task for wildlife conservation and provides 9 state-of-the-art swarm algorithms for benchmarking. All datasets, platforms, benchmarks, and video tutorials have been open-sourced and are available for public use. Our contributions can be summarized as follows:

\begin{table*}[t]
\caption{A detailed comparison of swarm UAV datasets. - indicates that specific information is not provided. DF: Discipline formation mode, where swarm UAVs keep a consistent and relatively static array; FF: Fixed formation mode, where each UAV navigates independently with a fixed path; AF: Autonomous formation mode, where each UAV flies autonomously. ET-Length: Each Trajectory Length. U2USim\textcolor{blue}{\textbf{$\star$}} represents the scalable U2USim.}
\label{table1}
\resizebox{\textwidth}{!}{%
    \begin{tabular}{c|ccccccccccc}
    \toprule 
    Dataset & Year & Terrains & Weather & Sensors & Formation & Real Data  & Tasks & Simulation & UAVs & ET-Lengh & Scalable \\
    \midrule
    CoPerception-UAVs    & 2022 & 1        & 1       & 1       & DF, FF          & -    & Basic     & AirSim + Carla & 5 & -        & N        \\
    CoPerception-UAVs+   & 2023 & 1        & 1       & 2       & DF, FF          & -    & Basic     & AirSim + Carla & 10   & -        & N        \\
    U2UData & 2024 & 4        & 7       & 8       & DF, FF, AF   & China   & Basic  & U2USim         & 3    & 15s      & N        \\
    \midrule
    \rowcolor[HTML]{ECF4FF} 
    {\bf U2UData+}               & 2025 & 4        & 7       & 8       & DF, FF, AF &  China   & \textbf{ELH}     & \textbf{U2USim\textcolor{blue}{$\star$}}  & \textbf{15}    & \textbf{600s}     & \textbf{Y}     \\
    \bottomrule
    \end{tabular}
}
\vspace{-0.3em}
\end{table*}

\begin{itemize}
	\item {\bf Dataset.} We collect the first swarm UAV autonomous flight dataset for ELH tasks, with a size of over 3.62T.
	\item {\bf Platform.} We build the first scalable swarm UAV data online collection and algorithm closed-loop verification platform, which allows users to collect customized datasets and verify algorithms.
	\item{\bf Benchmark.} We introduce an ELH task for wildlife conservation and provide comprehensive benchmarks with 9 SOTA models.
\end{itemize}

\section{Related Work}
This section introduces the related work of Swarm UAV autonomous flight methods, simulators and datasets in detail.

{\bf Swarm UAV autonomous flight.} Current low-altitude economy research mainly focuses on single-UAV autonomous flight and has matured core capabilities \cite{5, 13, feng2024multi}, including object detection, semantic segmentation, localization, obstacle avoidance, navigation, tracking, and stabilized flight control in controlled environments. However, they still suffer from many real-world challenges, for example: (1) Their perception remains fundamentally constrained by single-viewpoint occlusion and limited sensor range \cite{zheng2025cross, li2023lot}, severely reducing situational awareness in dynamic open environments. (2) Onboard computational resources restrict real-time decision-making for dynamic obstacle negotiation and executing ELH tasks \cite{feng2025embodied, li2025miv, shen2025detach}. (3) Operational robustness is inherently fragile \cite{gao2025vagu,gao2025suvad, gao2025evolution}, as hardware failures or unexpected obstacles often lead to task failure with no redundancy. Swarm UAV autonomous flight \cite{12} can solve the inherent limitations of single-UAV through collaborative perception, localization, communication, navigation, tracking, and task re-allocation. Due to the lack of datasets, research on autonomous flight algorithms for swarm UAV has just begun.

{\bf Swarm UAV simulators.} Existing swarm UAV simulators include FightGear \cite{S1}, XPlan \cite{S2}, Jmavsim \cite{S3}, Gazebo \cite{S4}, AirSim \cite{airsim}, Rfly-Sim \cite{S5}, Isaac Sim \cite{S6}, and U2USim \cite{u2usim}. Swarm UAV simulators need to more realistically simulate dynamic physical characteristics (such as collision); sensors such as IMU, camera, GPS, LiDAR, temperature, humidity, and airflow due to their small size; and interaction with the ROS ecosystem. FightGear is not open source. XPlan and Jmavsim can only interact with ROS. Gazebo, AirSim, and RflySim can interact with ROS, simulate physical collision, and output visual sensor content. AirSim and RflySim can also implement weather control. However the information on these simulators is purely simulated, and the models trained on these simulators are difficult to run in the real world. Isaac Sim and U2USim add real environment data based on previous simulators. Isaac Sim can visually realize digital twins of the real world through GPU rendering, but it is difficult to provide modal information other than visual and LiDAR modalities. U2USim is the first real-world mapping swarm UAV simulator, taking Yunnan Province as the prototype, including 4 terrains, 7 weather conditions, and 8 sensor types. However, all parameters of U2USim are fixed: it only contains 3 types of animals, the number of animals is fixed, the intensity and range of weather are fixed, and the take-off point of the UAV is also fixed. If we want to test in another terrain, we need to fly to the target location for a long time before each test.

{\bf Swarm UAV datasets.} Public swarm UAV datasets have significantly accelerated progress in UAV flight technologies in recent years. As shown in Table \ref{table1}, existing swarm UAV datasets include CoPerception-UAVs \cite{17}, CoPerception-UAVs+ \cite{36}, and U2UData \cite{feng2024u2udata}. CoPerception-UAVs \cite{17} and CoPerception-UAVs+ \cite{36} rely on open-source simulators like AirSim \cite{airsim} and CARLA \cite{CARLA}, featuring limited terrain, weather, and sensor types. These datasets collect data at fixed altitudes and in consistent or fixed formation modes. In contrast to autonomous driving \cite{liu2024survey}, UAVs' autonomous flight presents greater freedom, encounters more complex environments, and is more susceptible to the influence of temperature, humidity, and airflow due to its smaller size. Hence, there exists a notable domain gap between existing synthetic data and real-world data, potentially limiting the generalization of models trained. U2UData \cite{feng2024u2udata} is the first large-scale cooperative perception dataset for swarm UAVs autonomous flight, which is collected by three UAVs flying autonomously in the U2USim \cite{u2usim}, covering a 9 km$^2$ flight area, 4 terrains, 7 weather conditions, and 8 sensor types. U2UData manually selects 100 scenarios for each weather condition; U2UData collects 15 seconds of swarm UAV cooperative perception dataset for each scenario. Due to the emergence of U2UData, swarm UAV autonomous flight algorithms have begun to be studied. However, since U2UData only considers three UAVs tracking three animals, the length of each trajectory is only 15 seconds, and the dataset size and setting is fixed and cannot be expanded; only basic collaborative perception and tracking tasks can be designed. Complex ELH tasks \cite{liu2025tcdformer, Wang_improving} for swarm UAVs in dynamic open environments cannot be explored.

\begin{figure*}[t]
	\centering
	\subfigure[Simulator map]{
	\includegraphics[width=0.3\textwidth,height=0.2\textwidth]{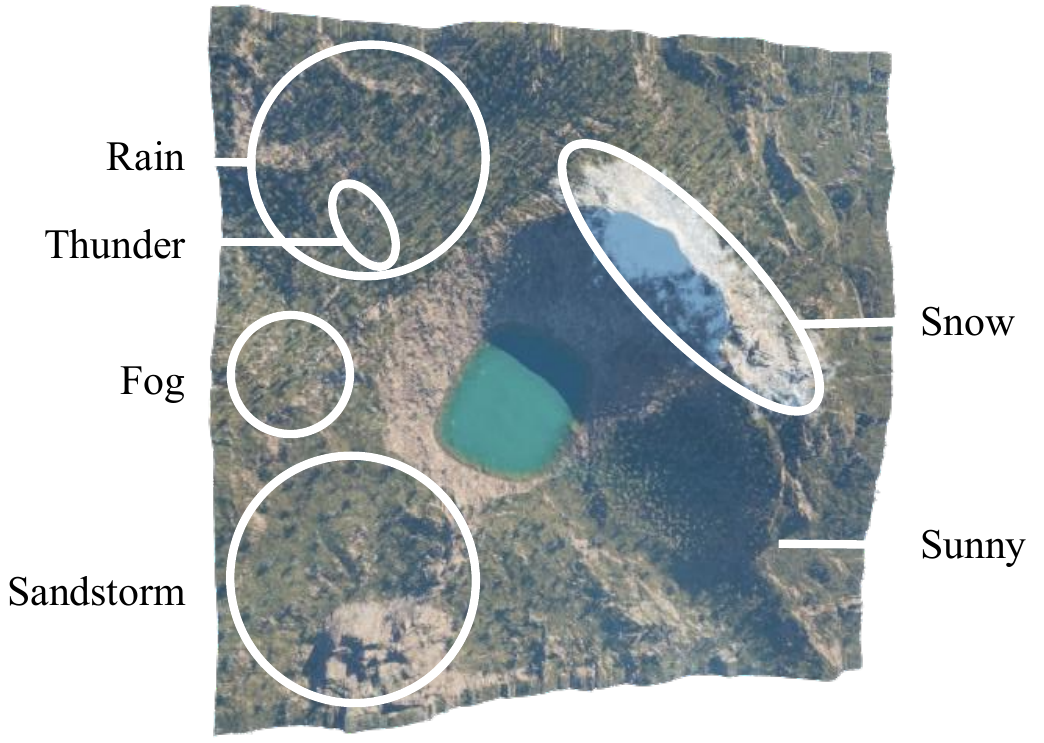}}
	\subfigure[Default starting point-weather-task]{
	\includegraphics[width=0.3\textwidth,height=0.2\textwidth]{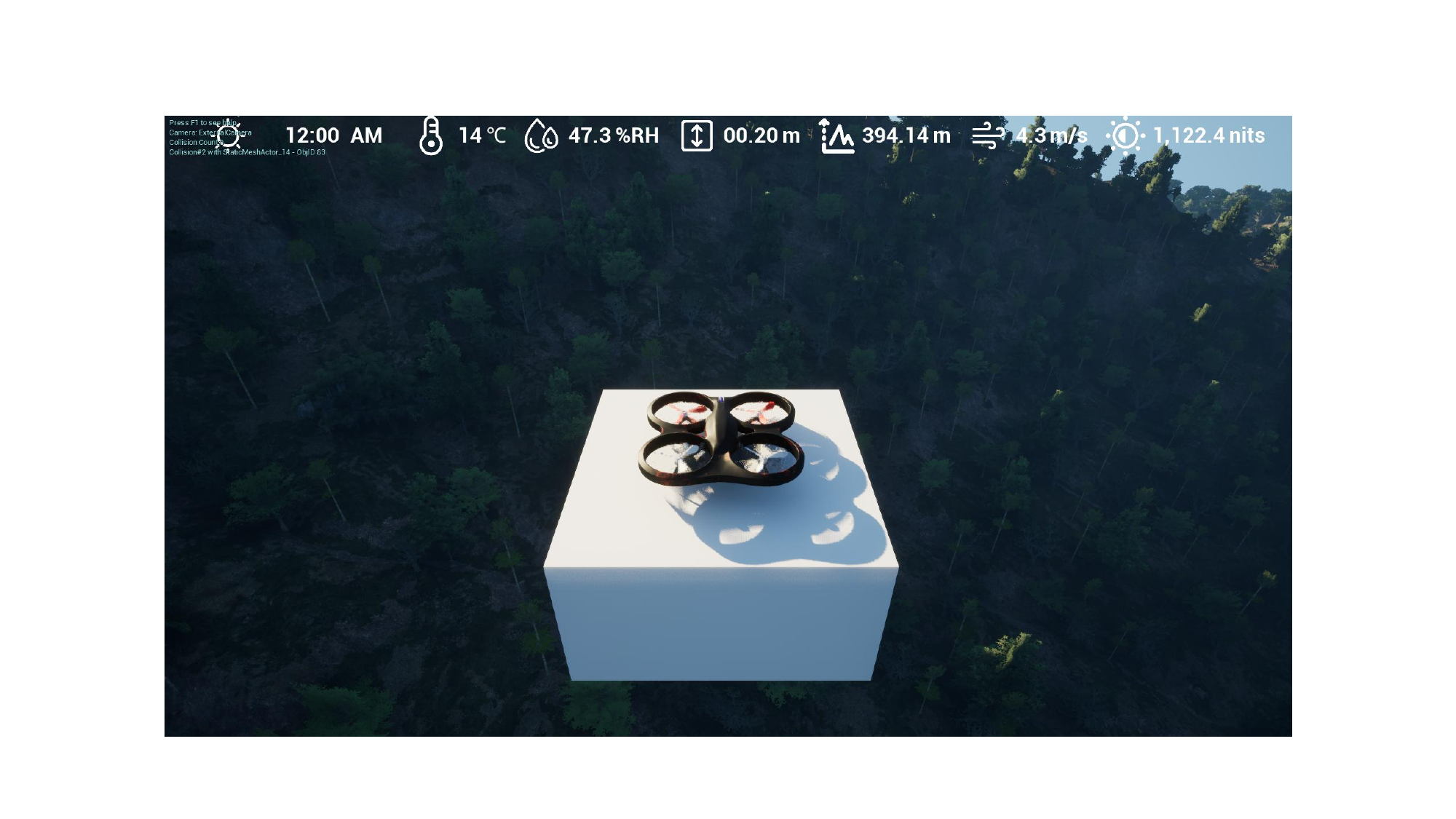}}
	\subfigure[Custom starting point-weather-task]{
	\includegraphics[width=0.36\textwidth,height=0.2\textwidth]{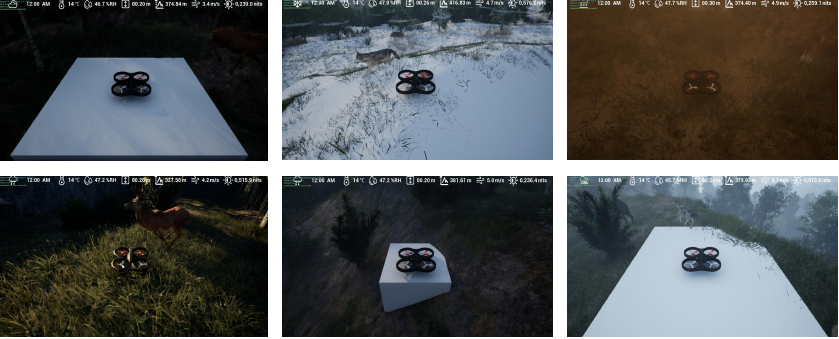}}
	\caption{The scalable UAVs data online collection and algorithm closed-loop verification platform. Wind throughout the map.}
	\label{fig_3}
	\vspace{-0.3em}
\end{figure*}

\section{U2UData+ Platform}
U2UData+ bulids the first scalable swarm UAV data online collection and algorithm closed-loop verification platform, as shown in Figure \ref{fig_1}, supporting the customization of simulators, UAVs, sensors, flight algorithms, formation modes, and ELH tasks. Through a visual control window, U2UData+ allows users to collect customized datasets through one-click deployment online and to verify algorithms by closed-loop simulation. We have built a {\it video tutorial} for this platform. For each scalable operation, users can complete it one by one according to the video tutorial.

{\bf Real-world mapping simulator.} The platform is based on U2USim \cite{u2usim}, a real-world mapping swarm UAV simulator. The platform uses Unreal Engine (UE) 5.2\footnote{https://www.unrealengine.com/en-US/unreal-engine-5} to construct a scaled-down 3km*3km simulated environment map based on the map of Yunnan Province. The platform includes 4 types of terrain: mountains, hills, plains and basins. The elevation range is [56.6, 3000]m. Based on the vegetation and animal distribution in Yunnan, 58 types of original forest vegetation and 15 types of animal assets were constructed, and more than 15 superposition methods were used to combine vegetation assets, including epiphytic growth, diagonal staggered growth, and so on. Among them, the leaves of each plant will dynamically change with wind, snow, and other weather conditions. This platform can represent complex ecological relationships between animals. It accurately models interactions such as predator-prey, showing how these relationships influence collective movement patterns. The platform includes 7 weather conditions: sunny, rain, snow, sandstorm, wind, thunder, and fog at specific positions within the simulation environment. The platform uses the real meteorological data of Yunnan Province collected by the China Meteorological Center to map the simulation environment based on longitude and latitude. Among them, temperature and humidity are scalars, and missing values are filled by the moving average method (interval 5m). Wind speed and wind direction are first decomposed into scalars along longitude and latitude, then missing values are filled by sliding average, and finally constructed by vector synthesis. 

{\bf Scalable simulator.} The simulator delivers extensive configurability through UE5.2, enabling dynamic adjustments to animal quantity and activity ranges, weather intensity and coverage, and UAV "starting point-weather-task" combinations. In the simulator startup interface, users can directly click the F11 key on the keyboard to make visual adjustments; input the animal quantity and the activity radius value; fine-tune the weather parameters using intuitive sliders, including intensity (e.g., rainfall severity, fog density) and spatial range. As shown in Figure \ref{fig_3}, six predefined UAV starting points are mapped to specific weather scenarios (rain, snow, sandstorm, thunder, fog, and sunny). Since wind is located throughout the map, there is no specific starting point setting. The starting point, weather, and task are added as options to the visual control window, and users can select from drop-down menus to implement custom "starting point-weather-task" combinations.

{\bf Scalable UAVs and sensors.} The platform includes 8 sensor types: RGB, depth, LiDAR, brightness, temperature, humidity, smoke, and airflow. These sensors are installed on the multirotor to explore the simulator map and collect data at 0.03-second intervals, which can be customized using a JSON settings file ("setting.json"). In this JSON file, UAV quantity is customizable; the type, quantity, position, angle, and resolution of sensors are also customizable, such as the Range and Number-Of-Channels of LiDAR sensors. The JSON file contains a total of 132 customizable parameters. Users can edit the JSON file to customize their own multirotor by selecting practical sensors and designing the sensor parameters. The specific meaning of each parameter and its modification range have been annotated in the open-source platform code.

\begin{figure}[t]
	\centering
	\includegraphics[width=0.48\textwidth]{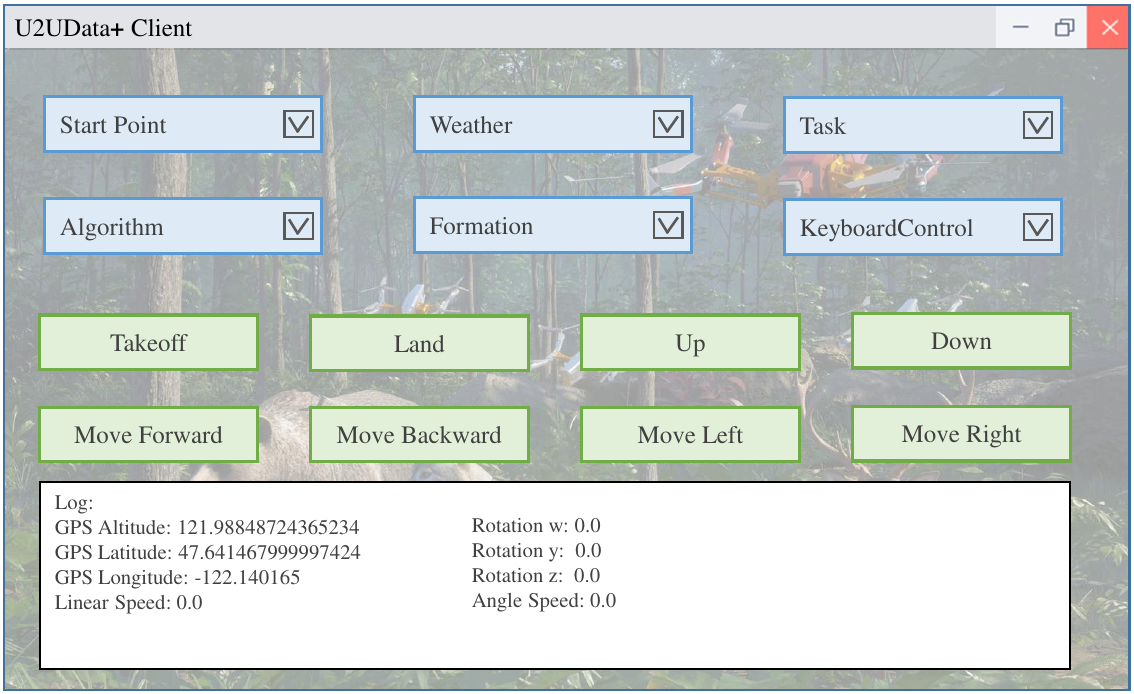}
	\caption{Visual control window. Users can collect customized datasets through one-click deployment online and verify algorithms by closed-loop simulation.}
	\label{fig_4}
    \vspace{-1.5em}
\end{figure}

\begin{table*}[t]
	\caption{A detailed comparison of the data size between U2UData+ with existing swarm UAV datasets.}
	\label{table2}
	\resizebox{\textwidth}{!}{%
		\begin{tabular}{c|ccccccccc}
			\toprule
			\multirow{2}{*}{Datasets} &
			\multicolumn{2}{c}{RGB} & 
			\multirow{2}{*}{Depth} &
			\multirow{2}{*}{LiDAR} & 
			\multicolumn{5}{c}{New Sensors} \\
			\cline{2-3}
			\cline{6-10}
			& RGB & Resolution & & & Airflow & Brightness & Temperature & Humidity & Smoke\\
			\midrule
			CoPerception-UAVs & 131.9K & 800*450 & - & - & - & - & - & - & - \\
			CoPerception-UAVs+ & 52.76K & 800*450 & - & - & - & - & - & - & - \\
			U2UData & 945K & 1920*1080 & 945K & 315K & 1.89M & 945K & 945K & 945K & 945K\\
			\midrule
			\rowcolor[HTML]{ECF4FF}
			U2UData+ & 12.96M & 1920*1080 & 12.96M & 4.32M & 25.92M & 12.96M & 12.96M & 12.96M & 12.96M\\
			\bottomrule                
		\end{tabular}%
	}
\end{table*}

{\bf Scalable ELH tasks.} The platform provides four preset ELH tasks: wildlife conservation employs adaptive animal tracking algorithms using real-time behavior prediction across variable terrains and vegetation density. Logistics distribution dynamically reroutes paths around simulated urban obstacles and weather disruptions while maintaining payload integrity. Patrol security implements anomaly detection through continuous environmental scanning, adapting surveillance patterns to emergent threats in real-time. Disaster rescue prioritizes survivor identification in volatile conditions (collapsing structures, spreading fires) via multi-sensor fusion and probabilistic hazard mapping. Each task integrates specialized perception-action loops that respond to unpredictable environmental changes without predefined waypoints, such as sudden weather shifts or moving obstacles. New ELH tasks (e.g., precision agriculture) can be added by modifying the UE5.2 simulator source code.

{\bf Scalable flight algorithms and formation modes.} The platform supports four swarm UAV autonomous flight algorithms for four ELH tasks: wildlife conservation, logistics distribution, patrol security, and disaster rescue. Those algorithms are built upon modularized components, including task planning, collaborative perception, localization, communication, navigation, tracking, and task re-allocation. Users can add or modify these algorithms via the open-source code of the visual control window, where modular code blocks allow drag-and-drop replacement or augmentation of existing logic. New autonomous flight algorithms for custom ELH tasks can be integrated by directly modifying the provided Python/ROS 2 interfaces in the code repository. The platform implements three distinct swarm formation modes: Discipline formation mode maintains strict geometric coordination (e.g., linear/radial arrays) for high-precision collaborative tasks, with real-time position correction compensating for environmental disturbances. Fixed formation mode enables individual UAVs to follow predefined paths, critical for infrastructure inspection or convoy protection scenarios. Autonomous formation mode supports dynamic reconfiguration where UAVs independently adapt spacing and topology using real-time perception data, ideal for complex environments like wildlife conservation or disaster rescue. Users can select any swarm formation modes via the visual control window for task-specific optimization.

{\bf Visual control window.} The platform provides a visual control window, as shown in Figure \ref{fig_4}. Users can collect datasets through a one-click deployment of the customized UAV starting point, weather, ELH tasks, swarm UAV autonomous flight algorithms, and swarm formation mode. Users can also verify swarm UAV autonomous flight algorithms by closed-loop simulation. For the platform basic capability test, users can first click the "keyboardControl" button, and then control the UAV by clicking the following buttons: "Take off", "Land", "Up", "Down", "Move Forward", "Move Backward", "Move Left", and "Move Right". We've also implemented XBOX controller control. Connect your XBOX and open the simulator to directly control the UAV with the controller. It's important to note that XBOX controller control and keyboard control are mutually exclusive.

\section{U2UData+ Dataset}
U2UData+ collects the first swarm UAV autonomous flight dataset for ELH tasks, with a size of over 3.62T. The dataset is collected by 15 UAVs in autonomous formation mode for the ELH task (wildlife conservation).

{\bf ELH tasks.} U2UData+ only collects one ELH task: wildlife conservation, with a size of over 3.62T. Due to the huge amount of data, users can collect datasets for other ELH tasks on the U2UData+ platform on their own.

\begin{table}[t]
\caption{Swarm UAV flight scene settings. ESTN: The trajectory number of each scene.}
\label{table3}
\centering
\resizebox{0.48\textwidth}{!}{%
\begin{tabular}{c|>{\columncolor[HTML]{ECF4FF}}c|c}
\toprule
Weather        & Scenes & ESTN \\
\midrule
Single-weather & \begin{tabular}[c]{@{}c@{}}Sunny, Rain, Snow, \\ Sandstorm, Thunder, Fog\end{tabular} & 5    \\
\midrule
Cross-weather  & \begin{tabular}[c]{@{}c@{}}Sunny-\textgreater{}Rain, Sunny-\textgreater{}Snow,\\ Sunny-\textgreater{}Fog, Sunny-\textgreater{}Sandstorm,\\ Rain-\textgreater{}Thunder, Rain-\textgreater{}Snow\end{tabular} & 3   \\
\bottomrule
\end{tabular}
}
\end{table}

\begin{table}[t]
\caption{Data collection settings between U2UData+ with existing swarm UAV datasets. ET-Length: The length of each trajectory. TNT: The total length of trajectories. }
\label{table4}
\resizebox{0.48\textwidth}{!}{%
\begin{tabular}{c|cccc}
\toprule
Datasets           & UAVs & Scenes & ET-Length & TLT \\
\midrule
CoPerception-UAVs  & 5    & 1      & -             & -         \\
CoPerception-UAVs+ & 10   & 1      & -             & -         \\
U2UData            & 3    & 7      & 15s           & 8.75h       \\
\midrule
\rowcolor[HTML]{ECF4FF}
U2UData+          & 15   & 12     & 600s       & 120h      \\
\bottomrule
\end{tabular}
}
\end{table}

{\bf Sensor setting.} The dataset bulids a comprehensive sensor suite including 5 RGBD cameras (1920x1080 resolution, 90° FOV, 30Hz sample rate), one 64-channel LiDAR (1 million points/second, 200m capturing range, ±3cm accuracy, -30° to 30° vertical FOV, -180° to 180° horizontal FOV, 10Hz sample rate), two airflow sensors measuring latitudinal and longitudinal wind speeds, and a GPS and IMU system providing odometry data. Complementary environmental sensors comprise one brightness sensor, one temperature sensor, one humidity sensor, and one smoke sensor. Navigation is enabled by integrated GPS and IMU systems providing odometry data. As shown in Figure \ref{fig_1}, all UAVs are equipped with 5 RGBD cameras (front, back, left, right, and bottom), a 64-LiDAR sensor (top), 1 brightness, temperature, humidity, and smoke sensor (bottom), 2 airflow sensors (back and right), and GPS/IMU systems. This multisensor configuration supports real-time environmental interaction across dynamic scenarios from LiDAR-based terrain mapping in the dense forest to airflow-adaptive flight control during storms. The synchronized RGBD cameras enable high-fidelity object tracking essential for wildlife monitoring.

\begin{figure*}[t]
	\centering
	\subfigure[Collaborative communication]{
	\includegraphics[width=0.32\textwidth]{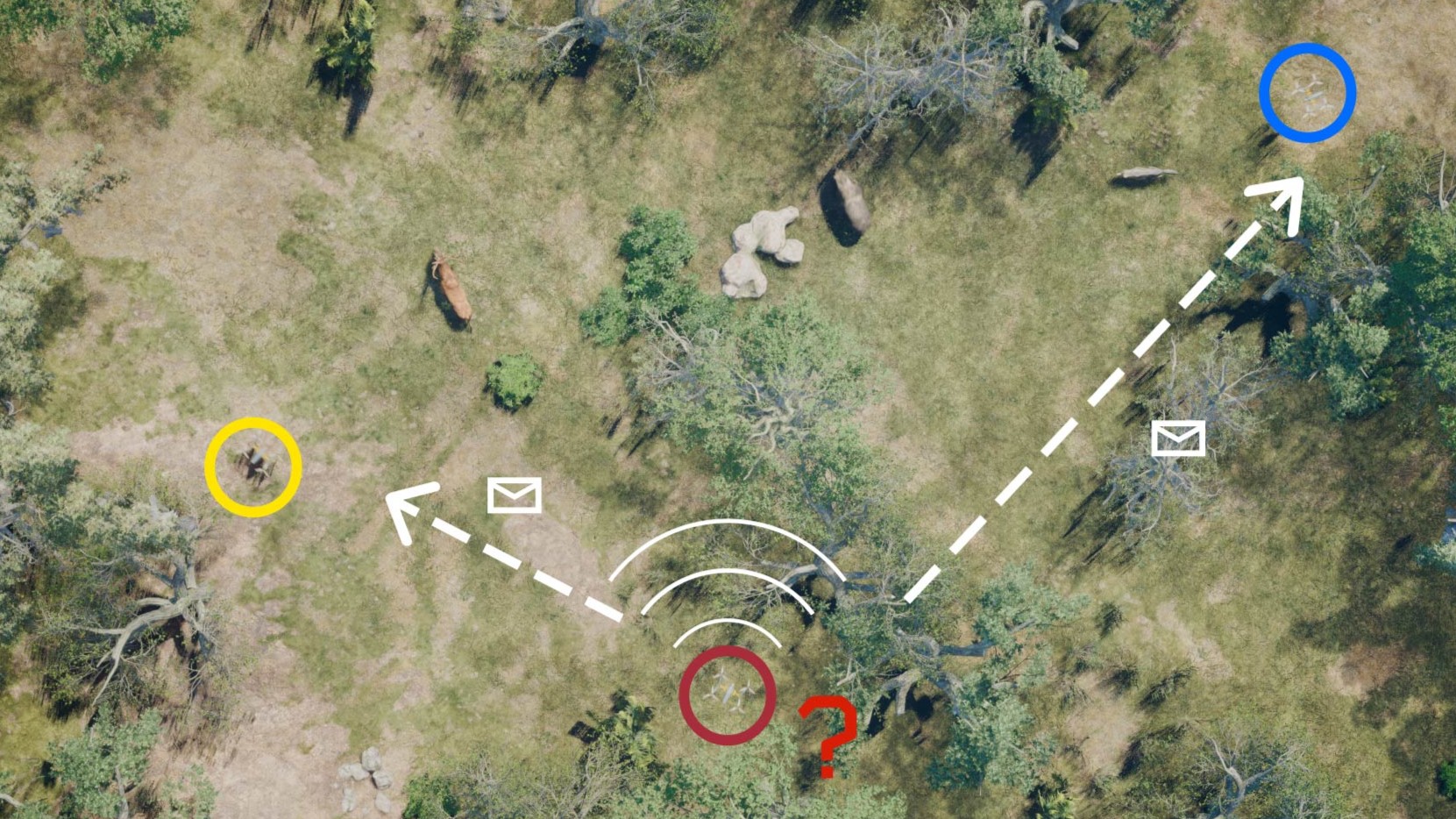}}
	\subfigure[Collaborative perception]{
	\includegraphics[width=0.32\textwidth]{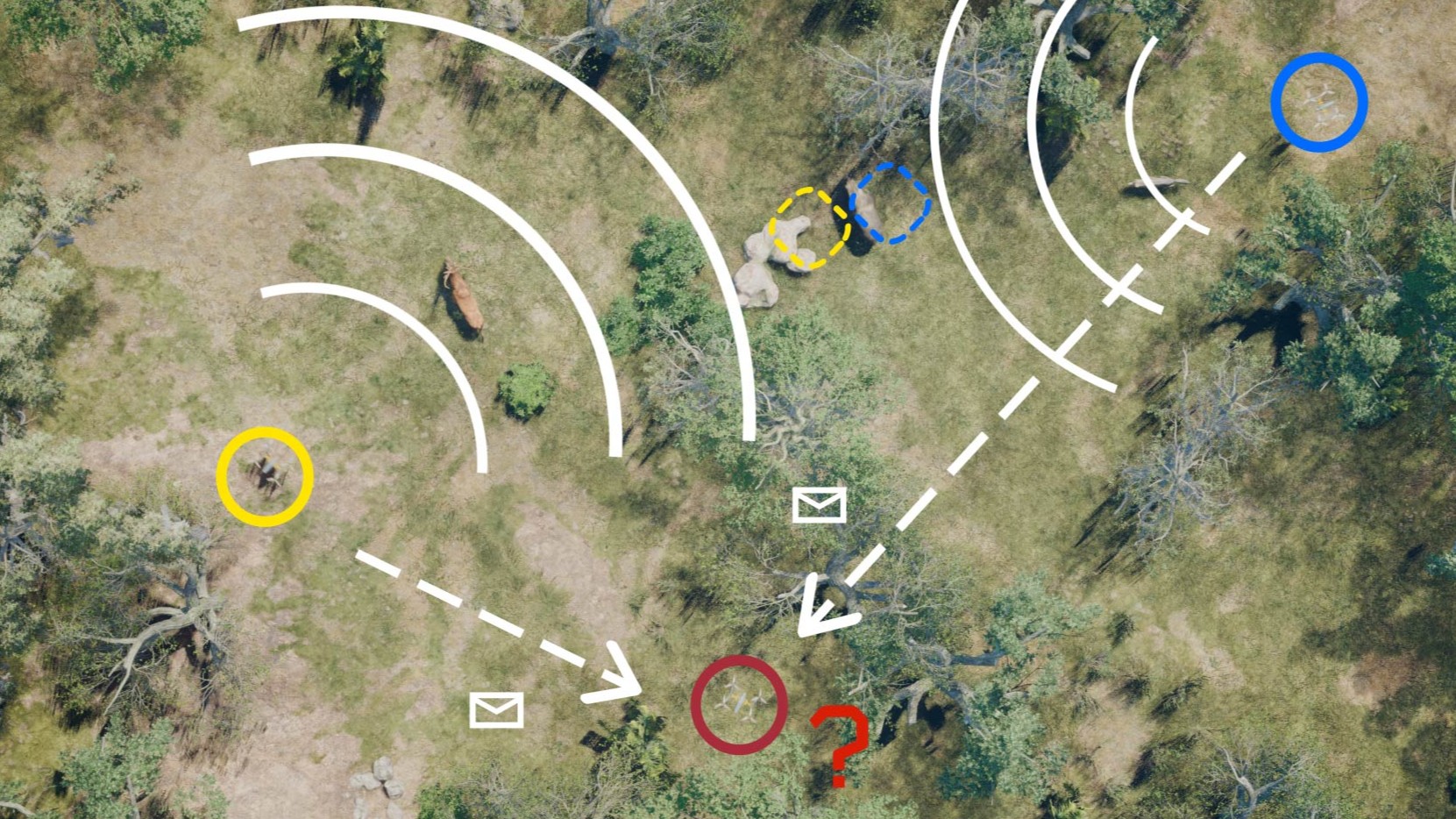}}
	\subfigure[Collaborative localization]{
	\includegraphics[width=0.32\textwidth]{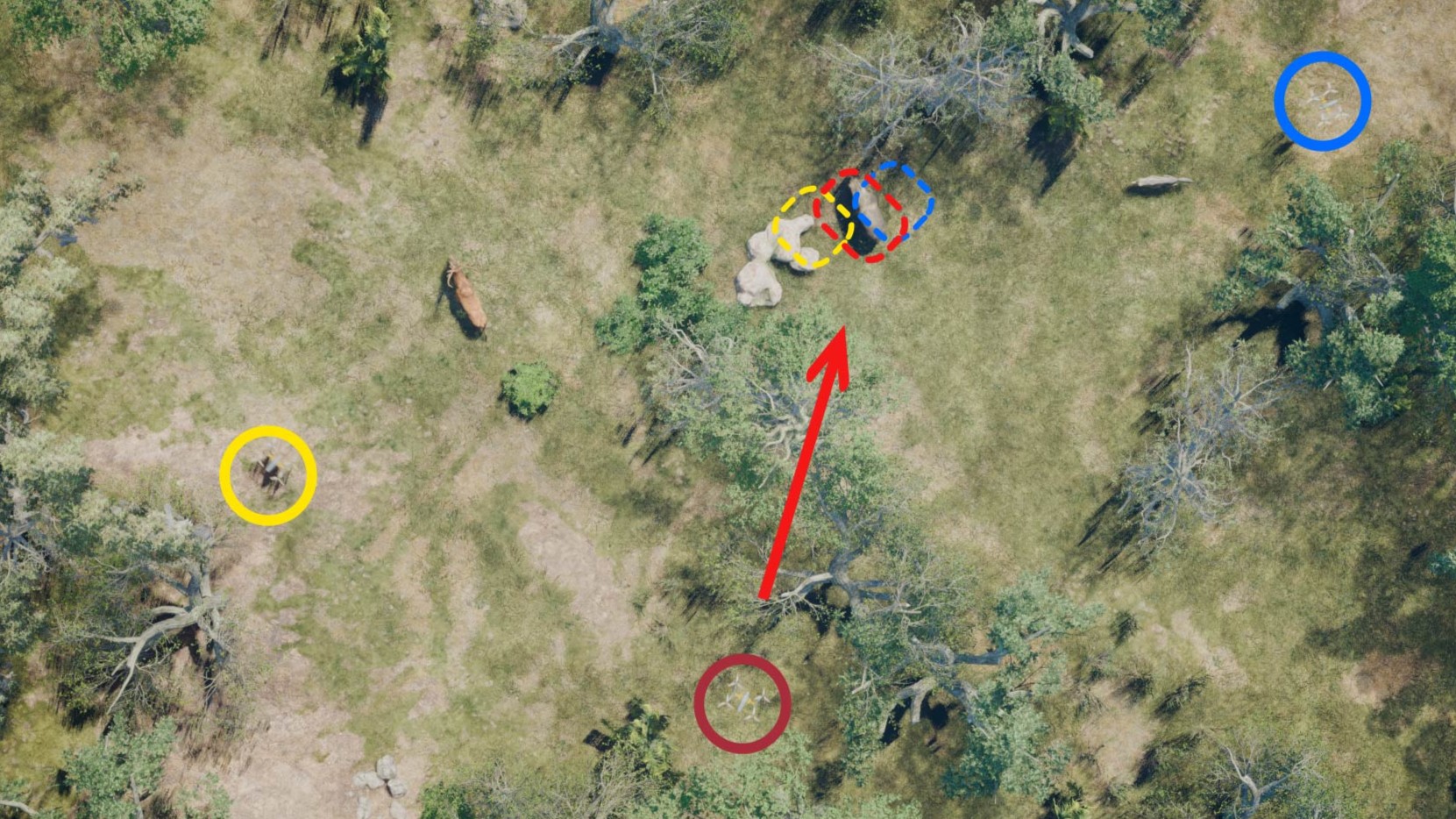}}
	
	\subfigure[Collaborative perception]{
	\includegraphics[width=0.32\textwidth]{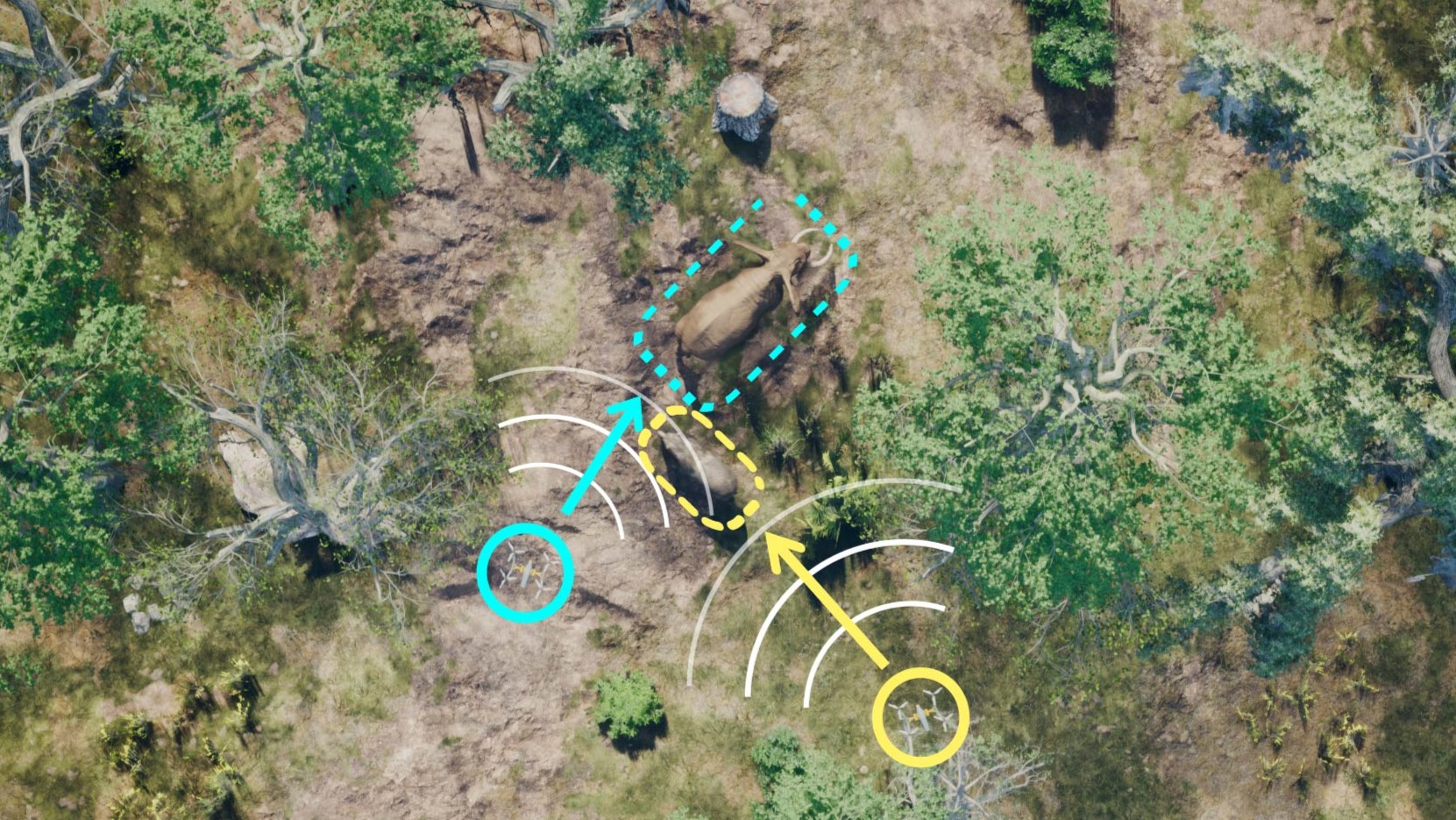}}
	\subfigure[Collaborative communication]{
	\includegraphics[width=0.32\textwidth]{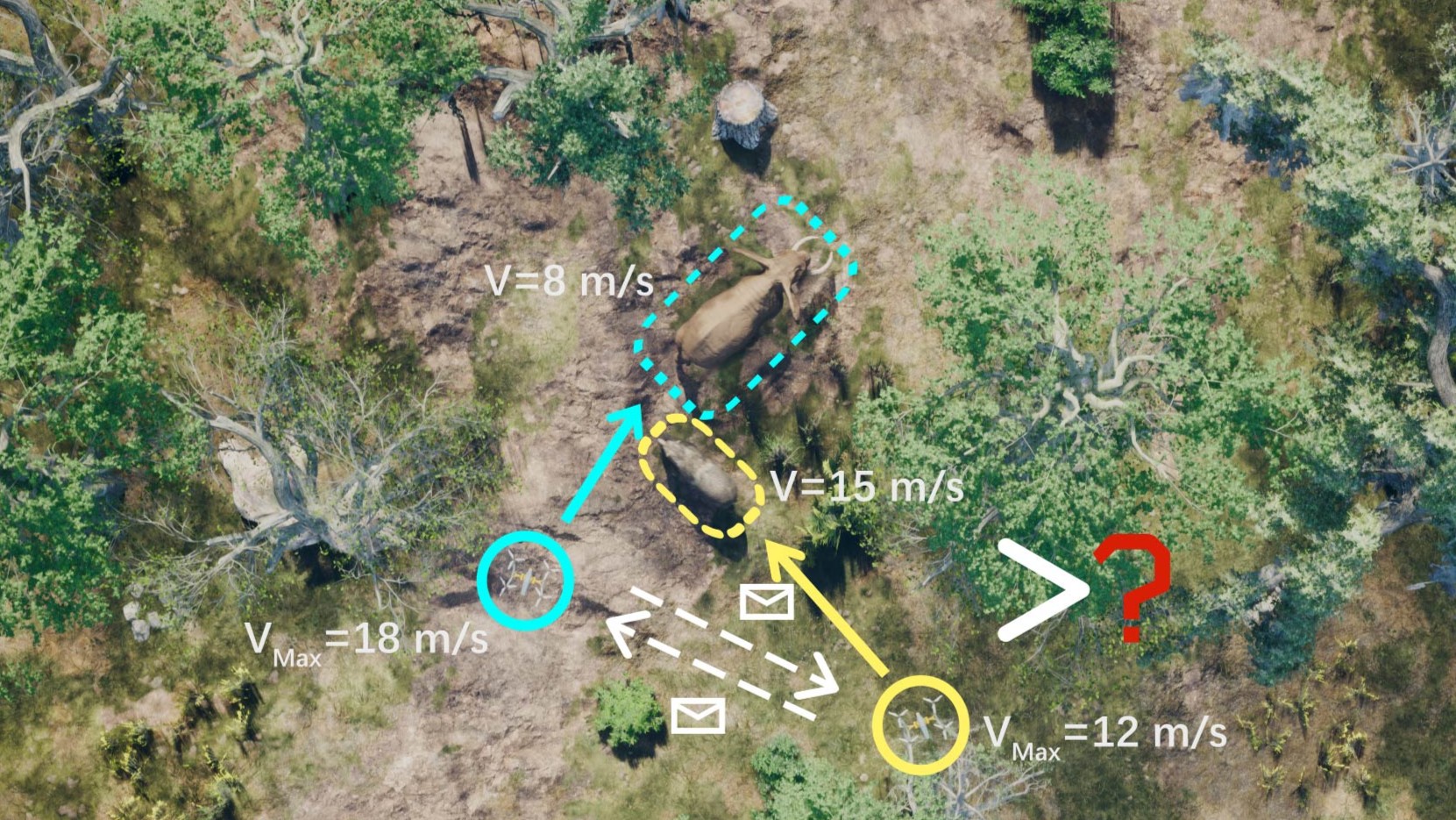}}
	\subfigure[Collaborative task re-allocation]{
	\includegraphics[width=0.32\textwidth]{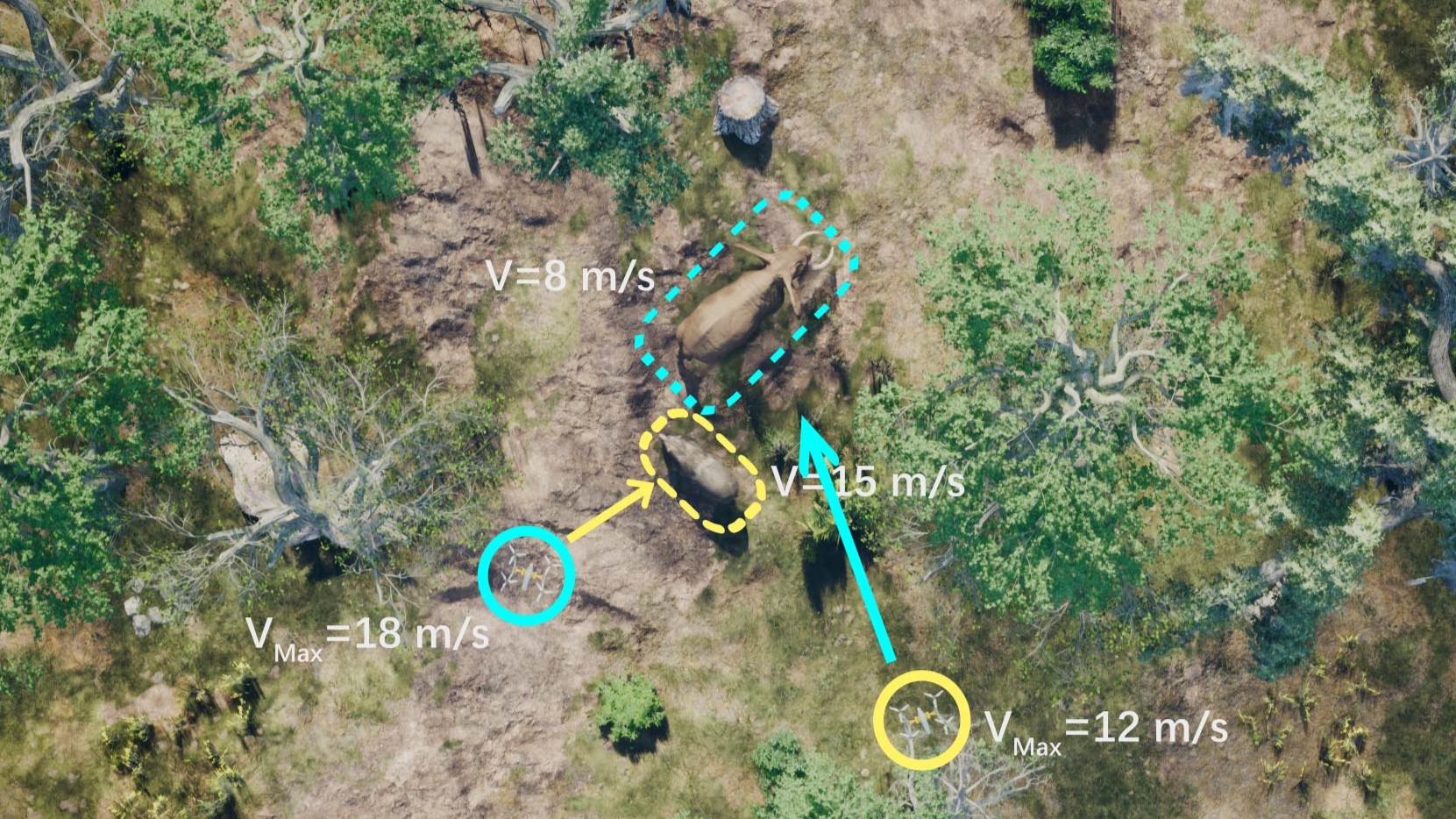}}
	\caption{The visualization of the U2UData+ dataset. we select two swarm UAV collaboration clips and annotate them.}
	\label{fig_5}
\end{figure*}

{\bf Scene setting.} The simulator map is first divided into 6 areas. Except for wind, which is located throughout the map, other weather is deployed in specific areas and has no intersection. For specific area locations, please watch the web page demonstration video. Since each specific area has different terrain, weather and terrain are strongly coupled. As shown in Table \ref{table3}, we construct 12 scenes based on the most common weather combinations. For single-weather scenes, the trajectory of each scene collected by U2UData+ is 5. For cross-weather scenes, the trajectory of each scene collected by U2UData+ is 3.

{\bf Dataset collection.} As shown in Table \ref{table4}, the dataset is collected by 15 UAVs in autonomous formation mode for the ELH task (wildlife conservation), comprising 12 scenes, 720 trajectories, and 600 seconds in length for each trajectory. The sampling interval of each sensor is 0.03 seconds and is synchronized in real time. As shown in Table \ref{table2}, we collect a total of 12.96M RGB frames, 12.96M depth frames, 4.32M LiDAR frames, 25.92M airflow frames, 12.96M brightness frames, 12.96M temperature frames, 12.96M humidity frames, and 12.96M smoke frames. The total length of the entire dataset is 120 hours. The total size of U2UData+ is 3.62T. The dataset has been open-sourced and are available for public use.

{\bf 3D bounding boxes annotation.} For annotating 3D bounding boxes on the gathered LiDAR data, we utilize SusTechPoint \cite{SusTechPoint}, a robust open-source labeling tool. There are a total of 15 object classes, and we annotate their 3D bounding box with 7 degrees of freedom, encompassing its location (x, y, z) and rotation (expressed as quaternions: w, x, y, z). The location (x, y, z) corresponds to the center of the bounding box. These 3D bounding boxes are annotated separately based on the global coordinate system of each UAV. This approach enables the sensor data from each UAV to be treated independently as a single-agent detection task. We initialize the relative pose of the two UAVs for each frame using positional information provided by the IMU on both UAVs.

{\bf Data usage.} We randomly divide the dataset into training sets, validation sets, and test sets according to the ratio of 0.7/0.15/0.15. It can greatly facilitate the credibility of the algorithm's performance compared to different papers.
\begin{table}[t]
\caption{A detailed comparison between U2UData and U2UData+. Basic tasks: collaborative perception and tracking. ELH tasks: wildlife conservation based on collaborative perception, localization, communication, navigation, tracking, and task re-allocation. \textbf{\textcolor{blue}{$\star$}} represents the scalable. \textbf{\textcolor{blue}{$\usym{2714}$}} represents the newly added function of U2UData+.}
\centering
\label{table5}
\begin{tabular}{c|c|>{\columncolor[HTML]{ECF4FF}}c}
\toprule
Comparison    & U2UData & U2UData+   \\
\midrule
Tasks                  & Basic tasks      & ELH tasks \textbf{\textcolor{blue}{$\star$}} \\
Each Trajectory Length & 15s              & 600s \textbf{\textcolor{blue}{$\star$}}              \\
ALL Trajectory Length  & 8.75h            & 120h \textbf{\textcolor{blue}{$\star$}}              \\
UAV Number             & 3                & 15 \textbf{\textcolor{blue}{$\star$}}                 \\
Tracking Goal          & 3                & 15 \textbf{\textcolor{blue}{$\star$}}                 \\
Sensor                 & 8                & 8 \textbf{\textcolor{blue}{$\star$}}                  \\
Flight Start Pointing  & Fixed            & Selected \textbf{\textcolor{blue}{$\star$}}            \\
Flight Algorithm       & Fixed            & Selected \textbf{\textcolor{blue}{$\star$}}           \\
Visual Control Window  & No               & \textbf{\textcolor{blue}{$\usym{2714}$}}                \\
Data online Collection & No               & \textbf{\textcolor{blue}{$\usym{2714}$}}                \\
Algorithm Closed-loop  & No               & \textbf{\textcolor{blue}{$\usym{2714}$}}               \\
\bottomrule
\end{tabular}
\end{table}

\begin{table*}[t]
	\caption{Swarm UAV collaborative tracking benchmark for ELH tasks in the U2UData+ dataset.}
	\label{table7}
	\centering
    \begin{tabular}{c|cccccc}
        \toprule
        Methods & AMOTA(↑) & AMOTP(↑) & sAMOTA(↑) & MOTA(↑) & MT(↑) & ML(↓) \\
        \midrule
        No Fusion & 9.36 & 25.48 & 32.19 & 23.47 & 18.67 & 65.52 \\
        Late Fusion & 14.62 & 31.68 & 47.96 & 37.41 & 27.93 & 37.28 \\
        Early Fusion & 18.61 & 32.45 & 43.80 & 41.64 & 25.51 & 34.48 \\
        When2Com & 20.16 & 34.32 & 49.47 & 45.74 & 30.69 & 32.51 \\
        DiscoNet & 20.94 & 37.56 & 52.63 & 46.79 & 32.50 & 29.47 \\
        V2VNet & 23.47 & 43.23 & \cellcolor[HTML]{ECF4FF}{\bf{57.82}} & 49.93 & \cellcolor[HTML]{ECF4FF}{\bf{35.68}} & 26.79 \\
        V2X-ViT & 22.86 & 40.76 & 55.74 & 48.70 & 33.26 & 27.94 \\
        CoBEVT & \cellcolor[HTML]{ECF4FF}{\bf{24.63}} & \cellcolor[HTML]{ECF4FF}{\bf{45.76}} & 54.73 & \cellcolor[HTML]{ECF4FF}{\bf{51.18}} & 34.79 & 27.25 \\
        Where2com & 24.16 & 42.63 & 55.69 & 50.82 & 33.76 & \cellcolor[HTML]{ECF4FF}{\bf{26.42}}\\
        \bottomrule
    \end{tabular}%
\end{table*}

{\bf U2UData+ vs. U2UData.} U2UData+ significantly expands upon its predecessor (U2UData) by transitioning from basic collaborative perception and tracking tasks to scalable ELH tasks (wildlife conservation) based on multi-UAV collaborative perception, localization, communication, navigation, tracking, and task re-allocation. As shown in Table \ref{table5}, key enhancements include 40× longer UAV trajectory of each scene (600s vs 15s) and 13.7× greater total trajectory duration (120h vs 8.75h), alongside 5× increases in UAV number (15 vs 3) and tracking targets number (15 vs 3). While retaining eight sensors per UAV, U2UData+ introduces dynamic flight algorithm selection and customizable starting points. Crucially, it adds three core innovations: a visual control window for real-time monitoring, one-click online data collection, and closed-loop algorithm validation. Those functionalities are absent in the original U2UData. Most importantly, U2UData+ establishes a scalable framework for swarm UAV autonomous flight in dynamic open environments, which can greatly alleviate the limitations of existing datasets on algorithm development.

{\bf Dataset Visualization.} U2UData+ dataset is the first large-scale swarm UAV autonomous flight dataset for the ELH task (wildlife conservation). As shown in Figure \ref{fig_5}, we select two swarm UAV collaboration clips and annotate them. The first clip ((a)-(c)) demonstrates that the target localization accuracy of a single UAV is limited due to obstacle obstruction and restricted field of view; swarm UAVs eliminate target localization errors through collaborative communication, perception, and localization. The second clip ((d)-(f)) illustrates that a single UAV makes it difficult to complete complex tasks independently due to its hardware limitations; swarm UAVs can improve the robustness of completing ELH tasks through collaborative perception, communication, and task re-allocation. These visualizations highlight the dataset's ability to design algorithms for ELH tasks.

\section{U2UData+ Benchmark}
{\bf SOTA Algorithms.} Since the algorithms of swarm UAV autonomous flight for ELH tasks are still lacking, we provide a swarm UAV collaborative tracking benchmark for ELH tasks in the U2UData+ dataset. This benchmark uses 9 SOAT collaborative tracking algorithms, including No Fusion, Late Fusion, Early Fusion, When2Com \cite{16}, DiscoNet \cite{15}, V2VNet \cite{12}, V2X-ViT \cite{14}, CoBEVT \cite{13}, and Where2com \cite{17}. This benchmark will be updated dynamically afterwards. 

{\bf Evaluation metrics.} We utilize the same evaluation metrics as outlined in \cite{tracking} for object tracking. These metrics include: AMOTA, average multiobject tracking accuracy; AMOTP, average multiobject tracking precision; sAMOTA, scaled average multiobject tracking accuracy, which ensures a more linear representation across the entire [0, 1] range of significantly challenging tracking tasks; MOTA, multi object tracking accuracy; MT, mostly tracked trajectories; ML, mostly lost trajectories.

{\bf Tracker.} We've chosen the AB3Dmot tracker \cite{tracking} as our basic module of all SOAT algorithms. This tracker initially retrieves 3D object detections from a LiDAR point cloud. It subsequently integrates the 3D Kalman filter with the birth and death memory technique to guarantee efficient and resilient tracking performance. It attains state-of-the-art performance while maintaining the fastest speed.

{\bf Implementation details.} We designate No Fusion as our baseline. To ensure a fair comparison, all models utilize PointPillar as the backbone for LiDAR feature extraction and use 32x feature compression (decompression) to save bandwidth. Among them, for CoBEVT, we only use the FuseBEVT module for feature aggregation without the SimBEVT module. During the training phase, we randomly designate one UAV as the ego UAV and train each model until achieving optimal task performance. During testing, we evaluate all compared models using a fixed ego UAV. For the tracking task, we utilize the previous three frames along with the current frame as inputs.

{\bf Results.} As shown in Table \ref{table7}, compared to the No Fusion method, AB3Dmot combined with cooperative algorithms significantly improves the tracking performance by at least 35.97{\%} AMOTA and 32.88{\%} sAMOTA. Compared with the Late Fusion method, the Intermediate Fusion method can improve the tracking performance by up to 27.48{\%} AMOTA. Compared with the Early Fusion method, the Intermediate Fusion method can improve the tracking performance up to 8.33{\%} AMOTA. 

\section{Conclusion}
Swarm UAV autonomous flight for ELH tasks is crucial. U2UData+ is the first large-scale swarm UAV autonomous flight dataset for ELH tasks and the first scalable swarm UAV data online collection and algorithm closed-loop verification platform. The dataset is captured by 15 UAVs in autonomous collaborative flight for ELH tasks, comprising 12 scenes, 720 traces, 120 hours, and 600 seconds per trajectory. The platform supports the customization of simulators, UAVs, sensors, flight algorithms, formation modes, and ELH tasks. Through a visual control window, this platform allows users to collect customized datasets through one-click deployment online and to verify algorithms by closed-loop simulation. U2UData+ also provides a benchmark with 9 SOTA models. In the future, we hope U2UData+ can assist swarm UAV algorithms in being deployed in the real world.

\section{Acknowledgments}
This work was supported by the National Natural Science Foundation of China (No. 62222209), Beijing National Research Center for Information Science and Technology under Grant No. BNR2023TD03006, China Postdoctoral Science Foundation under Grant No. 2024M751688, Postdoctoral Fellowship Program of CPSF under Grant No. GZC20240827, Open Research Fund Program of Beijing National Research Center for Information Science and Technology, and Beijing Key Lab of Networked Multimedia.

\bibliography{U2UData-2}

\end{document}